# Class-specific Differential Detection in Diffractive Optical Neural Networks Improves Inference Accuracy

*Jingxi Li[1,2,3†], Deniz Mengu[1,2,3†], Yi Luo[1,2,3], Yair Rivenson[1,2,3], Aydogan Ozcan[1,2,3*]*

[1]*Electrical and Computer Engineering Department, University of California, Los Angeles, CA, 90095, USA*
[2]*Bioengineering Department, University of California, Los Angeles, CA, 90095, USA*
[3]*California NanoSystems Institute, University of California, Los Angeles, CA, 90095, USA*
*†Equal contributing authors.*
*\*Corresponding author: ozcan@ucla.edu*

**Abstract**

Optical computing provides unique opportunities in terms of parallelization, scalability, power efficiency and computational speed, and has attracted major interest for machine learning. Diffractive deep neural networks have been introduced earlier as an optical machine learning framework that uses task-specific diffractive surfaces designed by deep learning to all-optically perform inference, achieving promising performance for object classification and imaging. Here we demonstrate systematic improvements in diffractive optical neural networks based on a differential measurement technique that mitigates the strict non-negativity constraint of light intensity. In this differential detection scheme, each class is assigned to a separate pair of detectors, behind a diffractive optical network, and the class inference is made by maximizing the normalized signal difference between the photodetector pairs. Using this differential detection scheme, involving 10 photodetector pairs behind 5 diffractive layers with a total of 0.2 Million neurons, we numerically achieved blind testing accuracies of 98.54%, 90.54% and 48.51% for MNIST, Fashion-MNIST and grayscale CIFAR-10 datasets, respectively. Moreover, by utilizing the inherent parallelization capability of optical systems, we reduced the cross-talk and optical signal coupling between the positive and negative detectors of each class by dividing the optical path into two jointly-trained diffractive neural networks that work in parallel. We further made use of this parallelization approach, and divided individual classes in a target dataset among multiple jointly-trained diffractive neural networks. Using this *class-specific differential detection* in jointly-optimized diffractive neural networks that operate in parallel, our simulations achieved blind testing accuracies of 98.52%, 91.48% and 50.82% for MNIST, Fashion-MNIST and grayscale CIFAR-10 datasets, respectively, coming close to the performance of some of the earlier generations of all-electronic deep neural networks, e.g., LeNet, which achieves classification accuracies of 98.77%, 90.27%, and 55.21% corresponding to the same datasets, respectively. In addition to these jointly-optimized diffractive neural networks, we also independently-optimized multiple diffractive networks and utilized them in a way that is similar to ensemble methods practiced in machine learning; using 3 independently-optimized differential diffractive neural networks that optically project their light onto a common output/detector plane, we numerically achieved blind testing accuracies of 98.59%, 91.06% and 51.44% for MNIST, Fashion-MNIST and grayscale CIFAR-10 datasets, respectively. Through these systematic advances in designing diffractive neural networks, the reported classification accuracies set the state-of-the-art for an all-optical neural network design, and the presented framework might be useful to bring optical neural network-based low power solutions for various machine learning applications and help us design new computational cameras that are task-specific.

**Introduction**

Machine learning, and in particular deep learning, has drastically impacted the area of information and data processing in recent years [1]–[5]. Research on optical machine learning has a very rich history [6]–[12], due to its advantages in terms of power efficiency, scalability, computational capacity and speed. With today's substantial computational power, advances in manufacturing technologies (e.g., 3D printing, lithography) and increasing availability of machine learning-related programming tools (e.g. TensorFlow), there has been remarkable progress on the use of machine learning in optics and photonics, focusing on e.g., the development of new integrated photonics devices [13]–[24] or the design of application specific free-space optical neural networks [25]–[27].

The task of object recognition and classification is an important application area of machine learning. It is conventionally realized in two main steps. First, a lens-based imaging system followed by a CMOS/CCD array captures a scene at hand. The digitized and stored image of the scene is then fed into an all-electronic artificial neural network (ANN) pre-trained for the task. The sampling density and, thus the number of detectors on the opto-electronic sensor plane, are dictated by the desired spatial and/or temporal resolution of the designed system [28]. In a classification system, high spatial resolution is generally desired due to the vital importance of spatial features for the performance of ANNs, forcing the pixel count and density of the sensor arrays to be relatively high, which, consequently, increases the requirements on the size of the memory as well as the computational power, inevitably hampering the effective frame-rate. Compressive sensing/sampling field has broadly aimed to overcome some of these resource inefficiencies in conventional optical systems. However, computationally demanding recovery algorithms associated with compressive sensing framework partially hinders its application for a wide range of areas in need of real-time operation.

In earlier work, we introduced diffractive deep neural networks [25], [27], which are composed of successive diffractive optical layers (transmissive and/or reflective), trained and designed using deep learning methods in a computer, after which it is physically fabricated to all-optically perform statistical inference based on its trained task at hand. In this framework, complex wave field of a given scene or object, illuminated by a coherent light source, propagates through the diffractive layers which collectively modulate the propagating light such that the intensity at the output plane of the diffractive network is distributed in a desired way, i.e., based on the specific classification or imaging task of interest, these diffractive layers jointly determine the output plane intensity in response to an input. The applications of this concept for the design of optical imaging systems as well as all-optical object classification were experimentally realized [25].

Unlike traditional, imaging-based machine vision systems, a diffractive optical neural network trained for a classification task needs only a few opto-electronic detectors, as many as the number of individual classes in a given dataset. Following their design and fabrication, diffractive optical neural networks execute classification with passive optical components, without the need for any power except the illumination beam and a simple *max* operation circuitry at the back-end. Unless optical nonlinearities are utilized, diffractive optical neural networks are linear in nature, except the final opto-electronic detector plane; despite its linearity, additional diffractive layers have been shown to improve the generalization and inference performance of the network, indicating the depth advantage that comes with the increasing number of diffractive neural layers in the optical network [25],[27]. With a single photo-detector assigned to each individual class of objects, Ref. [27] demonstrated a blind testing accuracy of 97.18% for all-optical classification of handwritten digits (MNIST database, where each digit was encoded in the amplitude channel of the input), and achieved 89.13% for all-optical classification of fashion-products (Fashion-MNIST database, where each object was encoded in the phase channel of the input).

In spite of the promising performance of the earlier work on diffractive optical networks, these architectures suffer from a well-known limitation in optics: the opto-electronic detectors are only sensitive to the incident optical power rather than the complex optical field, which limits the range of realizable values to non-negative real numbers. In this work, this non-negativity of the detected signal at the output plane of diffractive neural networks is mitigated through a differential detection scheme, which employs 2 opto-electronic detectors per data class at the output plane (see Fig. 1b). In this differential detection scheme, the output signal for each class is represented by the normalized difference of the signals of the corresponding detector pair. Therefore, half of the opto-electronic detectors are accounted for the positive part of the output signal, while the other half represents the negative part. The final inference of the diffractive optical network is simply made based on the maximum differential signal detected by these positive and negative detector pairs, each representing a separate class. Using this differential measurement scheme together with 5 diffractive layers having a total of 0.2 Million neurons, we numerically achieved blind testing accuracies of 98.54%, 90.54% and 48.51%, for MNIST, Fashion-MNIST and grayscale CIFAR-10 datasets, respectively (see Fig. 2 and Table 1). For comparison, without using the differential detection scheme, similar diffractive optical neural networks achieve blind testing accuracies of 97.51%, 89.85% and 45.20% for the same datasets, respectively.

In addition to the introduction of differential detection per class, in this work we also made use of parallel computation capability of passive diffractive layers, and jointly-optimized separate diffractive optical neural networks for positive and negative detectors (see e.g., Fig. 3), which are designed to work in parallel for differential inference of data classes. In some other implementations, we jointly-optimized a group of diffractive optical networks, where each diffractive network was specialized to infer a subset of classes (see e.g., Fig. 4). Our results demonstrate the significant advantages of these design strategies that use a combination of differential detection and class-specific optimization of individual diffractive neural networks that work all in parallel. For example, with *class-specific differential diffractive optical neural networks* that are jointly-optimized, we numerically achieved blind testing accuracies of 98.52%, 91.48% and 50.82% for MNIST, Fashion-MNIST and grayscale CIFAR-10 datasets, respectively, coming close to the inference performance of e.g., LeNet, an all-electronic deep neural network. Apart from these jointly-optimized diffractive neural networks, we also designed independently-optimized multiple diffractive neural networks, inspired by ensemble methods used in deep learning. For example, using 3 independently-optimized differential diffractive neural networks that separately project their diffracted light onto a common output/detector plane, we numerically achieved blind testing accuracies of 98.59%, 91.06% and 51.44% for MNIST, Fashion-MNIST and grayscale CIFAR-10 datasets, respectively.

Because of the passive nature of diffractive neural networks, at the cost of optical set-up alignment complexity as well as illumination power increase, one can create scalable, low-power and competitive solutions to perform optical computation and machine learning through these jointly-optimized diffractive neural network systems.

**Results and Discussion**

Figure 1 shows the examples of the input-output plane configurations for various diffractive neural network designs investigated and compared in this manuscript. For the sake of clarity, we devised a notation to represent these different optical network designs based on (1) the number of positive and negative detectors at the output plane, (2) the number of jointly-trained but independent diffractive networks (i.e., there is *no* electromagnetic coupling among individual diffractive networks), (3) the number of layers constituting each one of these individual diffractive neural networks, and (4) the number of neurons at each diffractive layer of an individual optical network. According to this notation, if a diffractive optical classifier has $Q^+$ positive detectors, $Q^-$ negative detectors at each output plane of $N$ jointly-optimized diffractive networks, each of which contains $L$ layers with $P$ neurons per layer, this optical classifier design is denoted by $D([Q^+,Q^-],[N,L,P])$. If the $Q^+$ positive detectors and $Q^-$ negative detectors are not at the same diffractive network output plane, i.e., decoupled from each other by distributing them to different optical neural networks, then the total number of jointly-optimized diffractive networks needs to be doubled to $2 \times N$, where each network will either have $Q^+$ or $Q^-$ detectors at the corresponding output plane. To reflect such a configuration, our notation is defined as $D([Q^+][Q^-],[2N,L,P])$ where the brackets separating the number of positive and negative detectors indicate that they are *not* placed on the same output plane, but rather they follow different diffractive optical networks that are all jointly-optimized, but operate individually without optical coupling from others. According to this notation, the standard diffractive optical neural network architectures (Fig. 1a) used in e.g., [25], [27] can be denoted as $D([M,0],[1,L,P])$ for a dataset with $M$ classes (e.g., $M = 10$ for MNIST). As another example, for a dataset with $M$ classes, a combination of class division and differential detection can lead to $2 \times M$ jointly-trained diffractive neural networks, with a *single* detector at the output plane of each diffractive network (corresponding to either positive or negative portion of a class signal), and this diffractive neural system is denoted, based on our notation, as: $D([1][1],[2M,L,P])$.

After the introduction of our notation to symbolize different diffractive neural systems ($D$), we now focus on quantifying the impact of some of these different designs on the inference and generalization performance of a diffractive classifier. First we start our analysis by comparing the performance of standard diffractive optical networks used earlier [25], [27], where there is a single optical network with $M = 10$ detectors at the output plane (e.g., $D([10,0],[1,5,40k])$), against the performance of a *differential* detection network with the same diffractive configuration, except the output plane, i.e., $D([10,10],[1,5,40k])$. Table 1, first and second rows reveal that maximizing the normalized differential signal for the target class improves the blind inference accuracy of a diffractive optical network (composed of 5 diffractive layers with a total of 0.2 Million neurons) to 98.54%, 90.54% and 48.51% for MNIST, Fashion-MNIST and grayscale CIFAR-10 datasets, respectively, compared to the corresponding accuracies achieved by $D([10,0],[1,5,40k])$, i.e., 97.51%, 89.85% and 45.20%, respectively. For a dataset with $M$ classes, this performance gain comes at the expense of a 2-fold increase in the number of opto-electronic detectors (*2M* instead of *M*), together with the use of additional but simple electronic read-out circuitry, composed of e.g., $M$ differential amplifiers and normalization logic; this extra computation at the output plane is rather straight-forward, with a computational complexity of $O(M)$.

When the optical path is divided into two as shown in Fig. 3a, we further increase the number of degrees of freedom of the diffractive neural system and decouple the optical signals detected by the positive and negative detector pairs; as a result of this, the blind testing accuracy of $D([10][10],[2,5,40k])$ shown in Fig. 3a increases further to 90.94% and 49.10% for Fashion-MNIST and CIFAR-10 datasets, respectively as shown in the last row of Table 1. For MNIST dataset, on the other hand, the average performance of *D([10][10],[2,5,40k])* remains approximately at the same level as *D([10,10],[1,5,40k])* architecture, i.e., 98.49% vs. 98.54%, respectively.

Table 2 summarizes our results on an alternative diffractive classifier design strategy: we jointly-trained a group of diffractive neural networks, where each one of them specialized on a sub-group of classes, and the opto-electronic detectors were placed at the output plane of the corresponding network. For example, as part of this design strategy, *D([2,0],[5,5,40k])* of Table 2 refers to a diffractive neural system that is composed of 5 jointly-trained diffractive neural networks, each having 5 diffractive layers (40k neurons per layer) and 2 detectors are placed at the corresponding output plane, where each detector represents one class of the dataset. Each one of these 5 diffractive neural networks is jointly-optimized together with the others, but does not have any optical coupling from the other networks. Based on our comparative analysis reported in Table 2 (non-differential row), the best performance among the non-differential diffractive designs is achieved when each diffractive optical network of a neural system specializes on *only one* class: *D([1,0],[10,5,40k])* achieved blind testing accuracies of 97.61%, 90.34% and 48.02%, for MNIST, Fashion-MNIST and grayscale CIFAR-10 datasets, respectively. The same conclusion regarding the success of class-specific diffractive neural networks also holds for differential detection strategy; Table 2 (differential rows) reports that *D([1,1],[10,5,40k])* and *D([1][1],[20,5,40k])* achieved the winner performance in this comparison for each differential row, with blind testing accuracies of 98.59% (98.52%), 91.37% (91.48%) and 50.09% (50.82%), for MNIST, Fashion-MNIST and grayscale CIFAR-10 datasets, respectively, where the values in parentheses refer to the performance of *D([1][1],[20,5,40k])*.

A direct comparison between the 'Differential' and 'Non-differential' rows of Table 2 further emphasizes the importance of the differential detection scheme. Not only the differential diffractive network designs show significantly better performance compared to their non-differential counterparts when the number of neurons and the number of diffractive layers are the same, but even *D([1,0],[10,5,40k])* architecture with 2 Million neurons in total cannot outperform *D([10,10],[1,5,40k])* that has 0.2 Million neurons in total, despite having 10 times more number of neurons in the diffractive classifier design.

Figure 5 summarizes the general conclusions that are revealed by our analysis: (**1**) differential diffractive neural systems outperform their non-differential counterparts; (**2**) diffractive neural networks that sub-specialize on a single class as part of a neural system outperform their counterparts that specialize on multiple classes per diffractive network; and (**3**) the combination of both design strategies, i.e., class-specific differential detection outperforms other counterpart diffractive neural system designs. These mean that for a dataset with *M* classes, *D([1][1],[2M,L,P])* would be a winner design, with *L* diffractive layers per class-specific optical network, and *P* neurons per diffractive layer (see Tables 2,3). For example, Table 2 reports that, for MNIST, Fashion-MNIST and grayscale CIFAR-10 datasets, *D([1][1],[2M=20,5,40k])* design achieves blind testing accuracies of 98.52%, 91.48% and 50.82%, respectively; to the best of our knowledge these values report the *highest accuracy* levels achieved so far for an all-optical neural network design.

So far, in our differential diffractive neural network designs, we considered balanced differential detection between the optical signals of $[Q^+]$ and $[Q^-]$, i.e., $[Q^+] - [Q^-]$. To further explore if this balanced differential detection is indeed an ideal choice, we also considered a more general case, where the two detectors of a pair assigned to a class can be merged with arbitrary scaling factors, $p_m$ and $n_m$, respectively (*m* represents the class number). We can generally denote this broader diffractive network design as: $D(p_m[M/N]\ n_m[M/N],[2N,L,P])$, where $p_m$ and $n_m$ can be any real number that can vary from class to class. For example, $p_m = n_m = 1$ refers to the standard balanced differential detection case considered so far, whereas $p_m = 1$, $n_m = -1$ case refers to a simple summation of the signals of the two detectors assigned to class *m*. By treating $p_m$ and $n_m$ as additional independent learnable parameters of a diffractive neural network, $D(p_m[M/N]\ n_m[M/N],[2N,L,P])$, we trained different designs that were initialized with random ($p_m$, $n_m$) values, which quickly converged to a solution with $p_m \approx n_m$ for each class of the corresponding dataset, proving empirically that a balanced differential detection is indeed preferred. We also noticed that the general design $D(p_m[M/N]\ n_m[M/N],[2N,L,P])$ with learnable detector coefficients did not improve our blind inference performance compared to the case of $p_m = n_m = 1$.

Another method to benefit from the parallel computing capability of passive diffractive neural networks is to create *independently-optimized* diffractive neural networks that optically project their diffracted light onto the same output/detector plane. Unlike the *jointly-optimized* diffractive neural systems described earlier, here in this alternative design strategy we select a diffractive network design, *D*, and independently optimize replicas of this design, where each network separately projects its diffracted pattern onto the same (i.e., common) detector plane. Not to interfere with the inference results of each diffractive neural network, here we considered intensity-only summation of the optical signals of each diffractive network at the common output plane, as opposed to coherent summation of the diffracted fields, which could perturb the predictions of each independent network due to constructive and destructive interference at the output plane. This can easily be achieved in a diffractive neural system by adjusting the relative optical path length differences between the individual diffractive networks to be larger than the temporal coherence length imposed by the bandwidth of the illumination source, ensuring that each detector at the common output plane sums up the optical intensities of all the individual diffractive neural networks. For each diffractive network of the ensemble, coherent operation is still maintained since the layer-to-layer separation in a given diffractive network is very small (e.g., 40λ for the designs considered in this work).

This design strategy of using independently-optimized diffractive networks is in fact similar to ensemble methods [29], [30] that are frequently used in machine learning literature. Figure 6 summarizes the blind testing accuracies achieved by this strategy using either non-differential or differential diffractive neural networks, i.e., *D([10,0],[1,5,40k])* or *D([10,10],[1,5,40k])*. For example, using 3 independently-optimized differential diffractive neural networks that optically project their light onto a common output plane with 10 detector pairs (one for each class), we numerically achieved blind testing accuracies of 98.59%, 91.06% and 51.44% for MNIST, Fashion-MNIST and grayscale CIFAR-10 datasets, respectively (see Fig. 6). Further increasing the number of independently-trained differential diffractive neural networks combined in an ensemble system brings diminishing return to the inference performance of the ensemble. For example, for CIFAR-10 dataset, optical classifier models that are composed of 2, 3 and 5 independently-optimized differential diffractive neural networks, *D([10,10],[1,5,40k])*, achieve blind testing accuracies of 50.68%, 51.44% and 51.82%, respectively. This diminishing return behavior stems from the increasing correlation between the output intensity distributions generated by the ensemble model and an additional independently-optimized diffractive neural network (to be added to the ensemble).

After reporting the results of various different design strategies for diffractive neural networks, in Table 3 we present a quantitative comparison of diffractive neural systems against some of the earlier hybrid (i.e., optical and electronic) neural networks as well as some of the widely-known all-electronic machine learning models used in the literature. This comparison once again highlights the importance of class-specific differential detection for improving the blind inference performance and the generalization of diffractive neural network systems. For example, *D([1][1],[20,5,40k])* matches the blind inference performance of convolutional deep neural networks such as LeNet and AlexNet for MNIST and Fashion-MNIST datasets, and falls short of the performance of LeNet for CIFAR-10 dataset only by 4.39%. A similar conclusion can be drawn from Table 3 for our comparison against the hybrid systems reported in [26], [27].

While the presented systematic advances in diffractive neural network designs have helped us achieve a competitive inference performance, with classification accuracies that are among the highest levels achieved so far for optical neural networks, there is still a considerable performance gap with respect to the state-of-the-art all-electronic deep learning models such as ResNet (see e.g., Table 3, CIFAR-10 performance comparisons). Despite its inferior performance compared to such all-electronic deep learning models that set the state-of-the-art in machine learning, class-specific differential diffractive neural networks still present several important advantages in terms of scalability, memory usage, computation speed and power efficiency since the main computation occurs all-optically and at the speed of light through diffraction within passive optical layers without the need for external power, except for the illumination light and a few detectors and related circuitry at the network output. Having underlined these important advantages, we should also note that significantly higher classification accuracies of state-of-the-art electronic deep neural networks such as ResNet once again emphasize the vital role of multi-channel convolutional layers and non-linearity inherent in these networks; as discussed in earlier work [25], [27], the use of nonlinear optical materials or optical resonances in diffractive neural networks can potentially improve the inference capabilities of diffractive neural systems beyond the currently presented results. Our results also reinforce an earlier conclusion regarding diffractive optical neural networks: their inference and generalization capabilities improve with additional diffractive layers jointly-designed and optimized by gradient-based learning, which illustrate the depth feature of diffractive neural systems, even if there is no non-linear optical material being employed per layer. Stated differently the general family of functions represented in this work through *D([M/N,M/N],[N,L,P])* or *D([M/N][M/N],[2N,L,P])* cannot be covered by a single diffractive optical layer, no matter how many neurons are employed.

Finally, we would like to also emphasize that these reported advances in the inference and generalization performance of class-specific differential diffractive neural networks come at the cost of a requirement to increase the input illumination power. For example, to keep the signal to noise ratio (SNR) of each photodetector that is positioned at an output plane of a class-specific differential diffractive neural network system, (e.g., *D([M/N,M/N],[N,L,P])*) at the same level as the SNR of the photodetectors of a standard diffractive neural network (i.e., *D([M,0],[1,L,P])*), the optical power of the input illumination beam must be increased by approximately *N* fold; the exact comparison is dataset and task dependent, and is actually governed by the photon efficiencies of different diffractive networks that make up of *D([M/N,M/N],[N,L,P])*. However, if *N* is increased to *M* (e.g., *M* = 10 for the datasets considered in this work), this means each diffractive network unit that is part of *D([M/N,M/N],[N,L,P])* has only 2 photodetectors at the corresponding output plane, whereas the standard diffractive neural network, *D([M,0],[1,L,P])*, has *M* = 10 photodetectors. Therefore, if we include in the training phase of the diffractive neural system a photon efficiency loss term for the photodetectors of *D([M/N,M/N],[N,L,P])*, penalizing poor diffraction efficiency per detector, one can potentially reduce this *N*-fold illumination power penalty by making class-specific networks more photon efficient as they deal with much smaller number of photodetectors at their output. To list another disadvantage of class-specific differential diffractive neural networks, because of their increased parallelism the complexity of the fabrication and alignment of the optical neural network set-up would be more complicated, and the overall size of the diffractive neural system would be increased compared to a single standard diffractive neural network. However, these are challenges that can be mitigated with 3D integrated photonic systems fabricated through e.g., lithography, and the need for increased optical illumination power is in general not a major concern due to various high-power lasers commonly available in different formats, including portable systems.

**Methods**

**Physical parameters of diffractive optical neural networks.** The physical model of wave propagation, used in the forward model of diffractive neural networks, was formulated based on the Rayleigh-Sommerfeld diffraction equation and digitally implemented, using a computer, based on the angular spectrum method [25]. According to this model, the neurons constituting the diffractive layers of an optical network can be interpreted as sources of modulated secondary waves [25]. Assuming an illumination wavelength of λ, each neuron provides an adequately wide diffraction cone enabling communication with all the neurons of the consecutive layer, provided that the size of each neuron is taken as ~0.5λ and the distances between the diffractive layers are set to be ~40λ. Diffractive optical neural networks designed based on these pre-determined (non-trainable) parameters are considered as fully-connected optical networks. All the diffractive optical neural networks presented in this manuscript were designed using this set of parameters (see Ref. [27] for further details and a comparison of design parameters of diffractive optical classifiers). In addition, the shape and size of each photodetector at a given output plane of a diffractive network were also fixed: we assumed square photodetectors, each with a width of 6.4λ. The form of the illumination, incident on the target objects, is assumed to be a uniform plane wave generated by a coherent light source and propagating parallel to the optical axis of the diffractive layers. According to our forward model, this incoming wave is modulated by an object at the input plane creating the complex wave field impinging on the 1st layer of a diffractive optical neural network after free-space propagation. The object functions of handwritten digits (MNIST dataset) were modeled as amplitude-only transmissive objects taking values between 0 (no transmission) and 1 (full transmission). The samples of Fashion-MNIST and CIFAR-10 datasets, on the other hand, were assumed to represent the phase channels of the transparent objects (unit amplitude transmission at every point), modulating only the phase of the input beam while preserving the amplitude distribution.

In our diffractive neural system and classifier designs, 5 fully-connected diffractive layers (phase-only modulation with each layer having 40k (200 × 200) neurons) were taken as *building blocks*. Although, our framework can be applied for the design of diffractive layers capable of modulating both the amplitude and phase of an incoming wave [27], fabrication of phase-only layers, in general, is preferable in terms of fabrication complexity and yield. Hence, the trainable parameter space for the diffractive optical classifiers investigated in this work contains only a phase modulation variable per neuron, resulting in a total of 0.2 million trainable variables for a 5-layer diffractive optical network, which constitutes the building block of the presented diffractive neural systems, *D*. For training of the diffractive optical neural networks reported in this work, the phase modulation parameter of each neuron was initialized as a Gaussian random variable with zero mean and 0.2π standard deviation.

**Implementation of differential diffractive optical neural networks.** Our differential detection model, in the context of diffractive optical classification systems, defines the class scores based on normalized differences between the positive and the negative detector signals at the output plane(s). With a pair of detectors assigned per class (a positive and a negative detector), the normalized difference for class *m*, is computed by:

$$I_{m,out} = \frac{I_{m,+} - I_{m,-}}{I_{m,+} + I_{m,-}}, \quad (1)$$

where, $I_{m,+}$ and $I_{m,-}$ stand for the optical signal of the positive and the negative detectors of class *m*, respectively. Due to scale-variant operation of the softmax function [31], the class scores ($I'_{m,out}$) were defined as the scaled versions of normalized differences in Eq. (1) according to;

$$I'_{m,out} = I_{m,out} / T, \quad (2)$$

where, *T* denotes a multiplicative scaling factor (also referred to as the 'temperature' hyperparameter in machine learning literature) and $I'_{m,out}$ is the class score of class *m*. For the results presented in this paper, *T* was set as 0.1, determined based on empirical observations. It is important to note that the sole purpose of Eq. 2 is to improve the speed of convergence of diffractive neural network optimization during the training phase and the blind testing classification performance of the final model [31], [32]. Therefore, when the softmax function is replaced with a *max* operation in the validation and testing processes, Eq. 2 is no longer used as part of our forward model and the blind prediction is solely made based on the output of Eq. (1).

The differential measurement technique is implemented using two different design approaches. In the first model, the positive and negative detectors representing a class are placed on the same output plane after a diffractive neural network, i.e., *D([M/N,M/N],[N,L,P])*. The second architecture, *D([M/N][M/N],[2N,L,P])*, is composed of 2*N* diffractive optical neural networks that independently control the light intensity detected by the positive and negative detectors assigned for different classes. Despite joint-optimization of the diffractive neural networks in these models, it was assumed that the diffractive networks are optically isolated from each other, meaning that the optical waves propagating through different diffractive neural networks do not interfere with each other.

Note that when *T* is set to be exponentially growing as a function of the number of epochs during the training phase, we observed a slightly better inference performance for *D([M/N,M/N],[N,L,P])* architecture. For example, in the case of *D([10,10],[1,5,40k])* when *T* was initialized as 0.1 and increased every 25 epochs by a multiplicative factor of *e* (e.g., at 50$^{th}$ epoch, $T = 0.1 \times e^2$), the blind testing accuracy achieved for CIFAR-10 dataset improved from 48.51% to 49.36%.

**Class-specific diffractive neural networks.** Division of elements of a target dataset into smaller sets based on their class labels was used to improve the inference performance of diffractive neural networks. In the training of class-specific diffractive neural networks, the target dataset was divided into sub-groups of classes and these sub-groups were split among parallel, simultaneously optimized diffractive neural networks. Although, these diffractive networks were trained simultaneously, the optical waves modulated by each network were assumed to be isolated from other diffractive networks of the same neural system, *D*. If used *without* the differential measurement scheme described earlier, the class-scores were directly calculated by the normalized signals of individual detectors placed at the output planes of the corresponding diffractive networks using:

$$I'_{m,out} = \frac{I_m}{\max(I_m) \times T} \quad (3)$$

where, $I_m$ denotes the optical signal of the detector assigned to class *m*, $\max(I_m)$ refers to the maximum optical signal among all the detectors, and *T* is a non-learnable hyper parameter used only during the training phase. For the designs presented in this work, *T* = 0.1 was selected empirically to improve the convergence speed and accuracy of the final model. As in the case of Eq. (2), once the joint-training of class-specific diffractive networks was completed, Eq. (3) was no longer used and the class predictions during the validation as well as blind testing stages were determined by selecting the *maximum* of the detected optical signals. When the class-specific diffractive neural networks were combined with the differential measurement method, Eq. (3) was accordingly replaced with the normalized signal difference calculation shown in Eq. (1) and the subsequent class score definition given in Eq. (2).

**Ensemble of diffractive optical neural networks.** Bagging [29] and ensemble [30], [33] methods are commonly used in machine learning literature to create multi-classifier systems that have superior performance compared to each individual unit constituting them. In these systems, the class scores coming from individual classifier units are merged into a single vector by means of arithmetic or geometric averaging, or by using majority voting schemes. Similarly, we used independently-optimized diffractive neural networks forming an ensemble and assumed that the diffracted optical signal from each optical network is super-imposed with the diffracted light of the other networks on the same (i.e., common) output plane, containing the photo-detectors. Assuming that the relative optical path length difference between any two diffractive networks of the ensemble is longer than the temporal coherence length of the illumination beam, the detectors at the output plane incoherently add up the light intensities generated by the independent diffractive networks. Apart from coherence engineering, an alternative option could be to sequentially measure the detector signals at the common output plane, one diffractive network at a given time, and digitally combine the class scores after the measurements. Both of these approaches (simultaneous incoherent summation of the projected light intensities at the common output plane vs. sequential capture of each diffractive network's output at the common detector plane and averaging of the class scores) achieved the same inference performance. To evaluate the performance of an ensemble of diffractive optical neural networks, we trained multiple replicas of a diffractive classifier design, *D*, by randomly changing the batch sequences and the initial phase modulation parameters of the diffractive layers for each replica. After every epoch, the corresponding model of each diffractive classifier unit was saved. When the training of all the individual units was finished, the best ensemble combination was selected based on their collaborative classification accuracy calculated using the validation dataset, and the blind classification accuracies provided by these best combinations on the *testing* dataset were presented under the Results and Discussion section.

The training strategy of setting *T* in Eq. (2) to be an exponentially growing parameter as a function of the number of epochs was also tested in the context of ensemble models. For example, a 3-unit ensemble model, where each individual differential diffractive network was trained using an exponentially growing *T*, achieved 50.86% blind testing accuracy, which is lower compared to 51.44% testing accuracy provided by the ensemble of 3 independently-optimized networks trained with a constant *T* = 0.1. A similar behavior was also observed for 2-unit ensemble models.

**Details of model training.** Object classification performances of all the models presented in this paper were trained and tested on three widely used datasets: MNIST, Fashion-MNIST and CIFAR-10. For MNIST and Fashion-MNIST datasets, 55000 samples were used as training data while the remaining 15000 objects were divided into two sets of 5000 and 10000 for validation and testing, respectively. The CIFAR-10 dataset was partitioned into three sets of 45000, 5000 and 10000 samples, used for training, validation and testing of our diffractive neural networks, respectively. Since the samples of CIFAR-10 dataset contain three color channels (red, green and blue), they were converted to grayscale using built-in *rgb_to_grayscale* function in TensorFlow to comply with the monochromatic (or quasi-monochromatic) illumination used in our diffractive network models.

Softmax cross-entropy was used as the loss function for all the neural network models (optical or electronic) presented in this work. With $I'_{m,\text{out}}$ denoting the class score of $m^{th}$ class, the classification loss can be computed by:

$$\text{Loss} = -\sum_{m=1}^{M} g_m \log(c_m) \quad (4)$$

where *M*, $c_m$ and $g_m$ denote (1) the total number classes in a given dataset, (2) the probability of an input being a member of class *m* according to softmax function, $\frac{\exp(I'_{m,\text{out}})}{\sum_{m=1}^{M} \exp(I'_{m,out})}$, and (3) the $m^{th}$ entry of the ground truth label vector, respectively.

All the neural networks in this paper (optical or electronic) were simulated using Python (v3.6.5) and Google TensorFlow (v1.10.0) framework. Adam optimizer was used [34] during the training of all models. The parameters of Adam optimizer were kept identical between each model, and taken as the default values in the TensorFlow implementation. The learning rate was initially set as 0.001, but an exponential decay was applied in every 8 epochs such that the new learning rate equals to 0.7 times the previous one. All the models were trained for 50 epochs and the best model was selected based on the classification performance on the validation set. For each model, the training was independently repeated 6 times with random batch sequences and initial phase modulation variables. Throughout this manuscript, our blind testing accuracy for each diffractive neural network design reports the mean value over these 6 repetitions, applied to testing datasets. For the training of our models, we used a desktop computer with a NVIDIA GeForce GTX 1080 Ti Graphical Processing Unit (GPU) and Intel Core (TM) i7-7700 CPU @3.60GHz and 16GB of RAM, running Microsoft Windows 10 operating system. The typical training time of the diffractive neural network shown in Fig. 2a is ~6 hours. For computationally more demanding architectures such as *D([1,1],[10,5,40k])* and *D([1][1],[20,5,40k])* the training time increased to ~26 hours and ~46 hours, respectively.

## Acknowledgements

Ozcan Research Lab acknowledges the funding of the Koç Group.

**List of Figures**

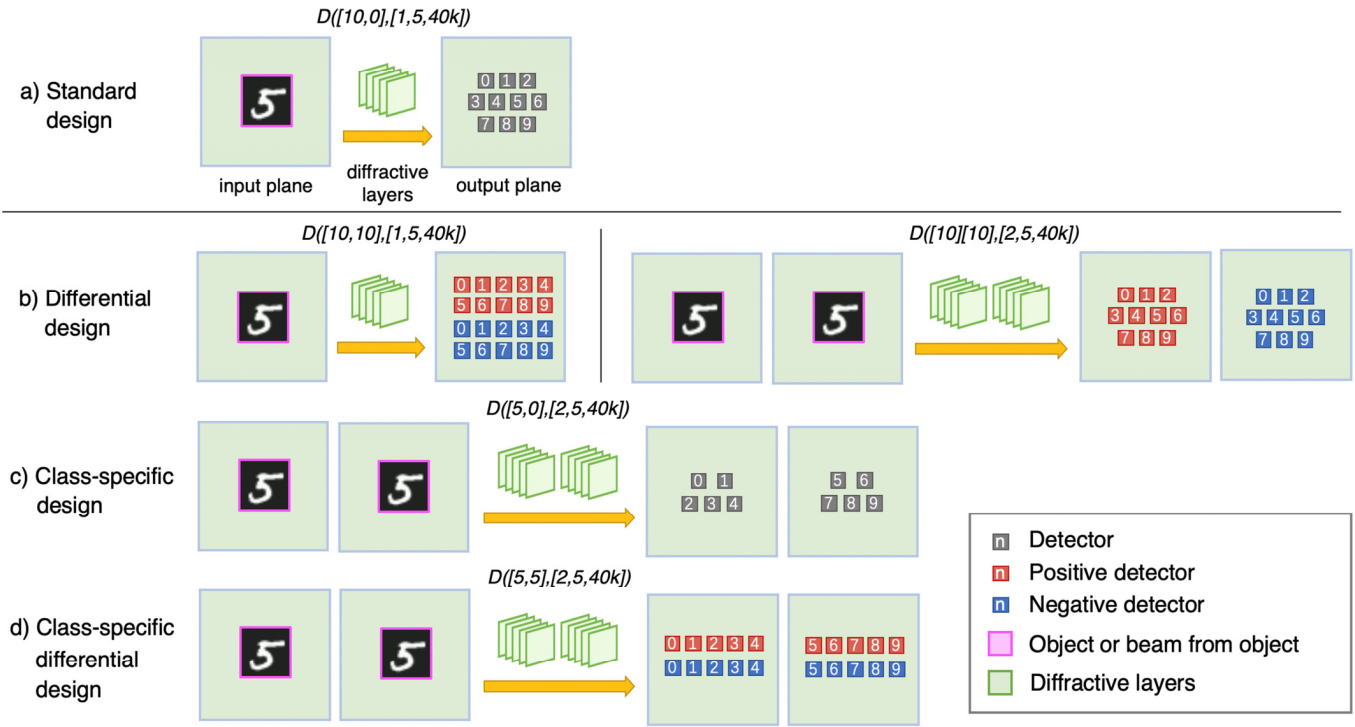

**Figure 1.** Illustration of different diffractive neural network design strategies. (a) Standard design refers to *D([M,0],[1,L,P])*, where *M* is the number of classes in the target dataset, which in this specific design is also equal to the number of detectors per diffractive neural network, *L* is the number of diffractive layers per optical network, and *P* refers to the number of neurons per diffractive layer. In the examples shown in this figure, *L*=5, *P*=40k, meaning 0.2 Million neurons in total. (b) Differential design shown on the left refers to *D([M,M],[1,L,P])*, whereas the one on the right refers to *D([M][M],[2,L,P])* as it uses two different jointly-optimized diffractive networks, separating the positive and the negative detectors by placing them at different output planes without optical coupling between the two. (c) Class-specific design shown here refers to *D([M/N,0],[N,L,P])*, where *N* > 1 is the number of class sub-sets (in this example, *N*=2 case is shown). (d) Class-specific differential design shown here refers to *D([M/N,M/N],[N,L,P])* where *N*=2 is illustrated. In general, there can be another version of a class-specific differential design where each diffractive neural network has only positive or negative detectors at the corresponding output plane; this special case is denoted with *D([M/N][M/N],[2N,L,P])*, where *2N* > 2 refers to the number of jointly-designed diffractive neural networks. *N*=1 case, i.e., *D([M][M],[2,L,P])* is included as part of (b) right panel, and we do not consider it under the class-specific neural network design since there is no class separation at the output/detector planes.

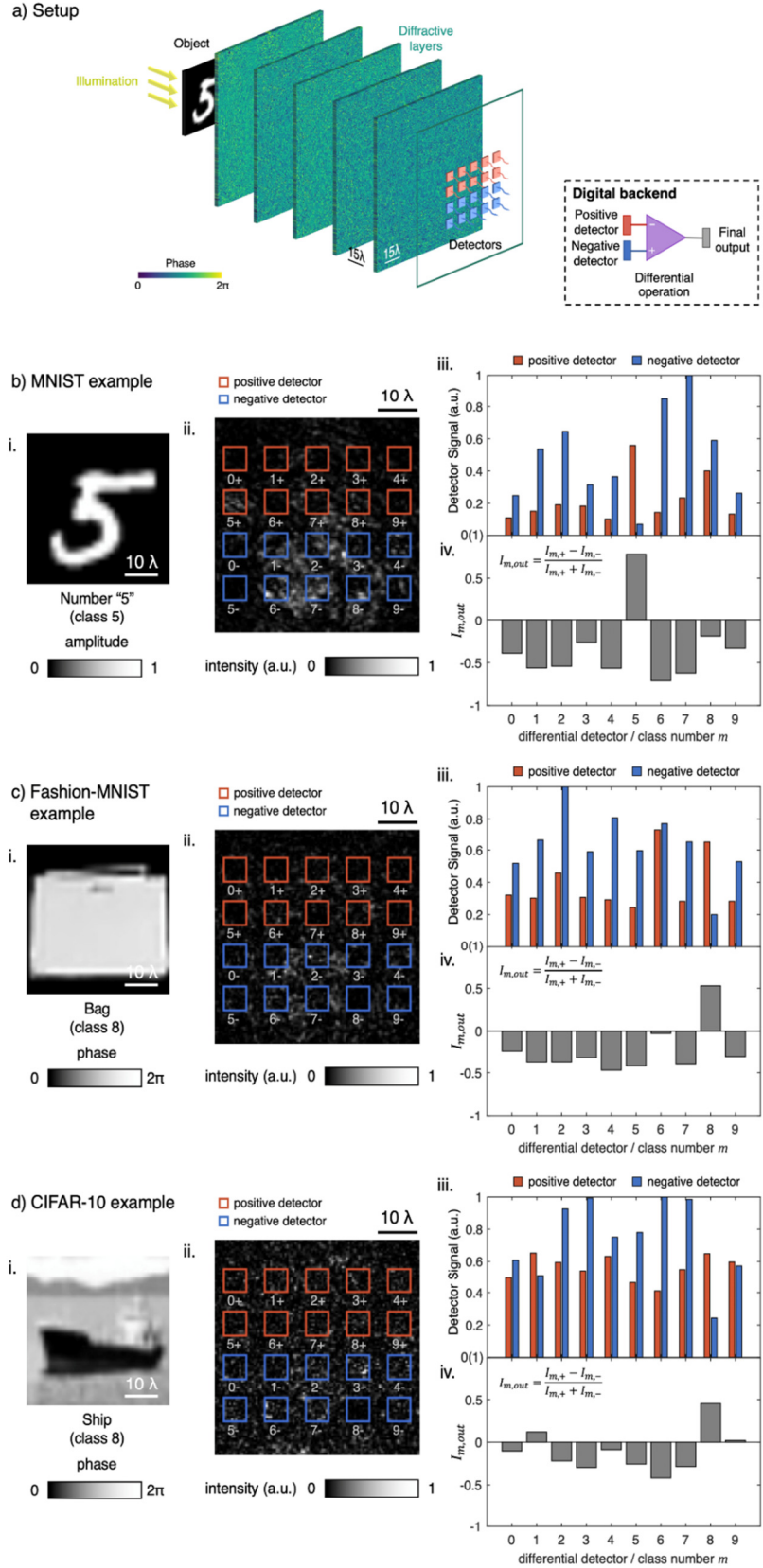

**Figure 2.** Operation principles of a differential diffractive optical neural network. (a) Setup of the differential design, *D([M,M],[1,L,P])*. In the example shown in this figure, *M*=10, *L*=5, *P*=40k. (b) A correctly classified test object from the
10

MNIST dataset is shown. Subparts of (b) illustrate the following: i. Target object placed at the input plane and illuminated by a uniform plane wave, ii. Normalized intensity distribution observed at the output plane of the diffractive optical neural network, iii. Normalized optical signal detected by the positive (red) and the negative (blue) detectors, iv. Differential class scores computed according Eq. (1) using the values in (iii). (c) and (d) are same as in (b), except for Fashion-MNIST and CIFAR-10 datasets, respectively. Note that while the input object in (b) is modeled as an amplitude-encoded object, the gray levels shown in (c) and (d) represent phase-encoded perfectly transparent input objects. Since diffractive optical neural networks operate using coherent illumination, phase and/or amplitude channels of the input plane can be used to represent information.

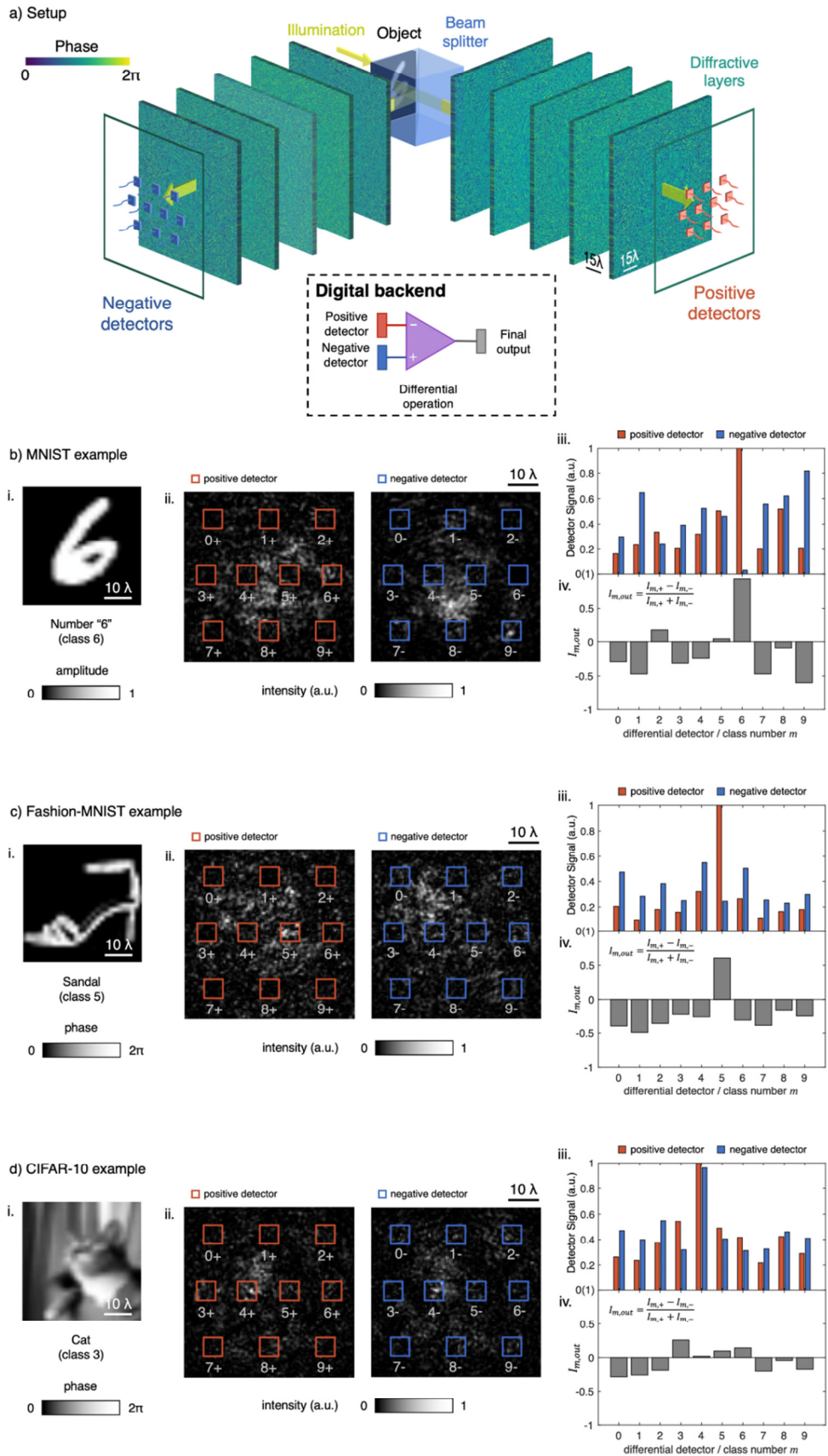

**Figure 3.** Operation principles of a diffractive optical neural network using differential detection scheme, where the positive and the negative detectors are split into two jointly-optimized networks based on their sign. (a) Setup of the differential design,

*D([M][M],[2,L,P])*. In the example shown in this figure, *M*=10, *L*=5, *P*=40k. (b) A correctly classified test object from the MNIST dataset is shown. Subparts of (b) illustrate the following: i. Target object placed at the input plane and illuminated by a uniform plane wave, ii. Normalized intensity distribution observed at the output plane of the diffractive optical neural network, iii. Normalized optical signal detected by the positive (red) and the negative (blue) detectors, iv. Differential class scores computed according Eq. (1) using the values in (iii). (c) and (d) are same as in (b), except for Fashion-MNIST and CIFAR-10 datasets, respectively. Note that while the input object in (b) is modeled as an amplitude-encoded object, the gray levels shown in (c) and (d) represent phase-encoded perfectly transparent input objects.

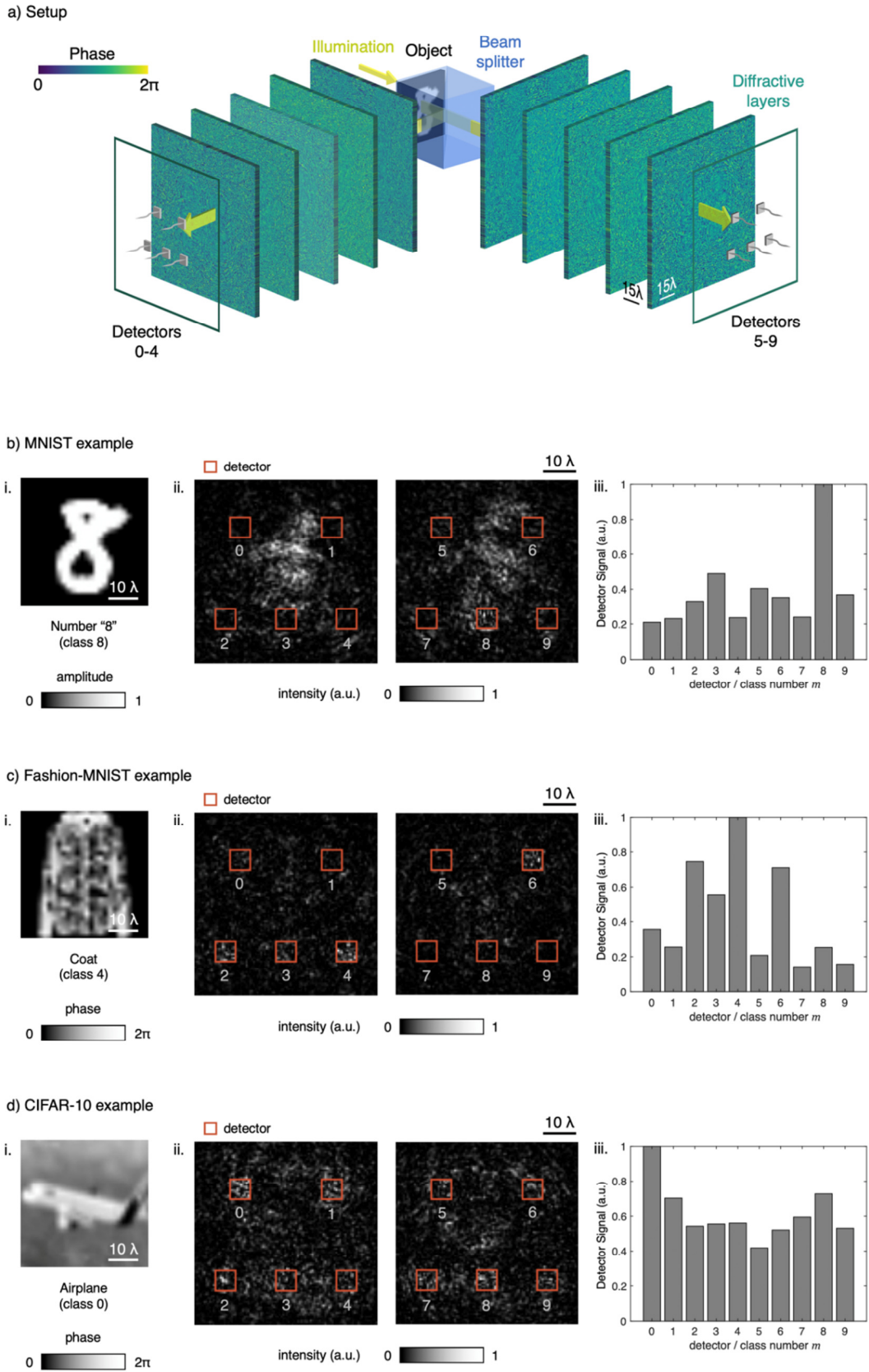

**Figure 4.** Operation principles of a diffractive optical neural network using class-specific detection scheme, where the individual class detectors are split into separate networks based on their classes. Unlike Figures 2 and 3 there are no negative

detectors in this design. (a) Setup of the class-specific design, $D([M/2,0],[2,L,P])$. In the example shown in this figure, $M$=10, $L$=5, $P$=40k. (b) A correctly classified test object from the MNIST dataset is shown. Subparts of (b) illustrate the following: i. Target object placed at the input plane and illuminated by a uniform plane wave, ii. Normalized intensity distribution observed at the two output planes of the diffractive optical neural networks, iii. Normalized optical signal detected by the detectors. (c) and (d) are same as in (b), except for Fashion-MNIST and CIFAR-10 datasets, respectively. Note that while the input object in (b) is modeled as an amplitude-encoded object, the gray levels shown in (c) and (d) represent phase-encoded perfectly transparent input objects.

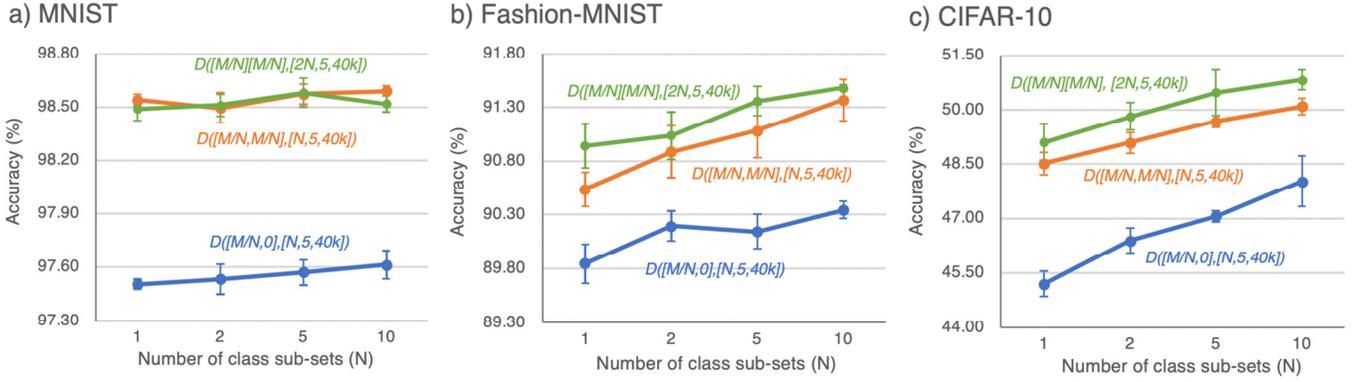

**Figure 5.** Performance comparison of different diffractive neural network systems as a function of *N*, the number of class sub-sets. *M* = 10 classes exist for each dataset: MNIST, Fashion MNIST and grayscale CIFAR-10. Based on our notation, *N=M*=10 refers to a jointly-optimized diffractive neural network system that specializes to each one of the classes separately. These results confirm that class-specific differential diffractive neural networks (*D([M/N][M/N],[2N,L,P])* for *N*>1) outperform other counterpart diffractive neural network designs. For each data point, the training of the corresponding diffractive optical neural network model was repeated 6 times with random initial phase modulation variables and random batch sequences; therefore each data point reflects the mean blind testing accuracy of these 6 trained networks, also showing the corresponding standard deviation.

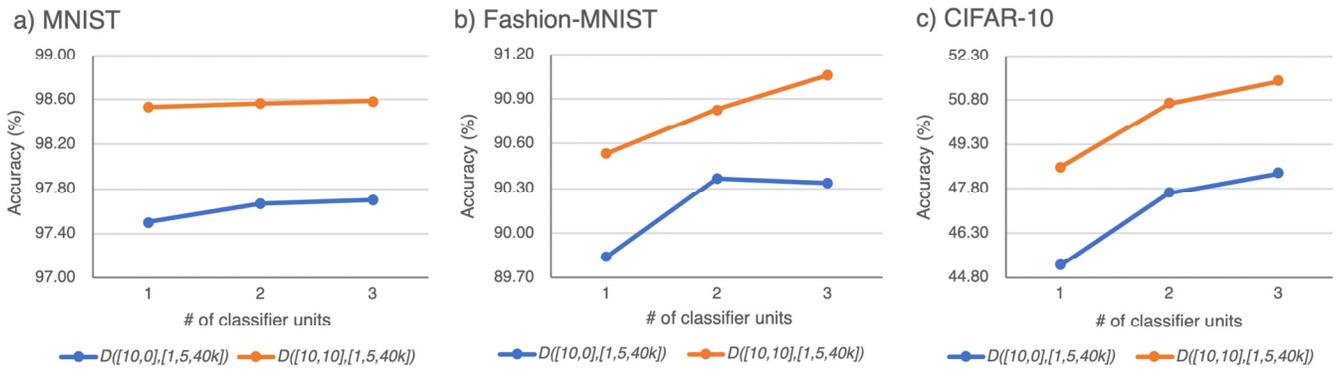

**Figure 6.** The comparison between the classification accuracies of ensemble models formed by 1, 2 and 3 independently-optimized diffractive neural networks that optically project their diffracted light onto the same output/detector plane. Blue and orange curves represent *D([10,0],[1,5,40k])* and *D([10,10],[1,5,40k])* designs, respectively. (a) MNIST, (b) Fashion-MNIST and (c) grayscale CIFAR-10. Not to perturb the inference results of each diffractive network due to constructive/destructive interference of light, incoherent summation of the optical signals of each diffractive network at the common output plane is considered here, which can be achieved by adjusting the relative optical path length differences between the individual diffractive networks to be larger than the temporal coherence length of the illumination source.

# List of Tables

| Architecture | MNIST | Fashion | CIFAR-10 (grayscale) |
|---|---|---|---|
| *D([10,0],[1,5,40k])* | 97.51 ± 0.03 | 89.85 ± 0.18 | 45.20 ± 0.35 |
| *D([10,10],[1,5,40k])* | 98.54 ± 0.03 | 90.54 ± 0.16 | 48.51 ± 0.30 |
| *D([10][10],[2,5,40k])* | 98.49 ± 0.03 | 90.94 ± 0.16 | 49.10 ± 0.30 |

**Table 1**. Blind testing classification accuracies of non-differential (top row) and differential diffractive optical networks, without any class specificity or division. $M = 10$ classes exist for each dataset: MNIST, Fashion MNIST and gray-scaled CIFAR-10. For each data point, the training of the corresponding diffractive optical neural network model was independently repeated 6 times with random initial phase modulation variables and random batch sequences; therefore each data point reflects the mean blind testing accuracy of these 6 trained networks, also showing the corresponding standard deviation.

| Type | Architecture | MNIST | Fashion | CIFAR-10 (grayscale) |
|---|---|---|---|---|
| Class-specific Non-differential *D([M/N,0],[N,L,P])* $N>1$ | *D([5,0],[2,5,40k])* | 97.53 ± 0.08 | 90.19 ± 0.14 | 46.37 ± 0.35 |
| | *D([2,0],[5,5,40k])* | 97.57 ± 0.07 | 90.14 ± 0.16 | 47.05 ± 0.16 |
| | *D([1,0],[10,5,40k))* | 97.61 ± 0.08 | 90.34 ± 0.08 | 48.02 ± 0.70 |
| Class-specific Differential *D([M/N,M/N],[N,L,P])* $N>1$ | *D([5,5],[2,5,40k])* | 98.50 ± 0.09 | 90.89 ± 0.24 | 49.09 ± 0.24 |
| | *D([2,2],[5,5,40k])* | 98.57 ±0.06 | 91.08 ± 0.25 | 49.68 ±0.17 |
| | *D([1,1],[10,5,40k])* | 98.59 ±0.03 | 91.37 ± 0.19 | 50.09 ± 0.23 |
| Class-specific Differential *D([M/N][M/N],[2N,L,P])* $N>1$ | *D([5][5],[4,5,40k])* | 98.51 ± 0.08 | 91.04 ± 0.22 | 49.82 ± 0.38 |
| | *D([2][2],[10,5,40k])* | 98.58 ± 0.06 | 91.36 ± 0.13 | 50.47 ± 0.63 |
| | *D([1][1],[20,5,40k])* | 98.52 ± 0.05 | 91.48 ± 0.03 | 50.82 ± 0.26 |

**Table 2**. Blind testing classification accuracies of different class division architectures combined with non-differential and differential diffractive neural network designs. For each data point, the training of the corresponding diffractive optical neural network model was independently repeated 6 times with random initial phase modulation variables and random batch sequences; therefore each data point reflects the mean blind testing accuracy of these 6 trained networks, also showing the corresponding standard deviation.

| Type | Network Architecture | MNIST (%) | Fashion (%) | CIFAR-10 (%) |
|---|---|---|---|---|
| Optical (Diffractive) | Standard design *D([10,0],[1,5,40k])* | 97.51 ± 0.03 | 89.85 ± 0.18 | 45.20 ± 0.35 |
| | Differential design *D([10,10],[1,5,40k])* | 98.54 ± 0.03 | 90.54 ± 0.16 | 48.51 ± 0.30 |
| | Ensemble of 3 differential designs *D([10,10],[1,5,40k])* | 98.59 | 91.06 | 51.44 |
| | Class-specific differential design *D([1,1],[10,5,40k])* | 98.59 ± 0.03 | 91.37 ± 0.19 | 50.24 ± 0.17 |
| | Class-specific differential design *D([1][1],[20,5,40k])* | 98.52 ± 0.05 | 91.48 ± 0.03 | 50.82 ± 0.26 |
| Hybrid (Optical + Electronic) | Ref. [27] | 98.97 | 90.45 | - |
| | Ref. [26] | - | - | 51.00 ± 1.40 |
| Electronic | SVM [35] | 91.90 | 83.20 | 37.13 |
| | LeNet [36] | 98.77 | 90.27 | 55.21 |
| | AlexNet [2] | 99.20 | 89.90 | 72.14 |
| | ResNet [37] | 99.51 | 93.23 | 88.78 |

**Table 3**. Comparison of blind testing accuracies of different types of neural networks, including Optical, Hybrid and Electronic.